\documentclass{article} 
\usepackage{iclr2023_conference,times}
\usepackage{graphicx}

\usepackage{tikz}
\usepackage{comment}
\usepackage{amsmath,amssymb} 
\usepackage{stmaryrd}
\usepackage{color}
\usepackage{mathrsfs}

\usepackage[T1]{fontenc}
\usepackage{url}
\usepackage{booktabs}       
\usepackage[pagebackref=true,colorlinks=true,citecolor=blue,linkcolor=blue]{hyperref}        
\usepackage{amsfonts}       
\usepackage{graphicx}
\usepackage{subcaption}
\usepackage{xspace}
\usepackage{bm}
\usepackage{multirow}
\usepackage{wrapfig}
\usepackage{tabu}
\usepackage{sidecap, caption}
\usepackage{colortbl}
\usepackage[capitalise]{cleveref}
\usepackage{footnote}
\usepackage{algorithm}
\usepackage{algpseudocode}
\usepackage{wrapfig}
\usepackage{authblk}
\makesavenoteenv{tabular}
\newcommand{\ubold}[1]{\fontseries{b}\selectfont#1}
\renewcommand\_{\leavevmode \kern.06em\vbox{\hrule width.3em}}
\definecolor{Gray}{gray}{0.85}
\def\eg{e.g.\xspace}
\def\ie{i.e.\xspace}
\definecolor{MediumBlue}{HTML}{0000CD}
\definecolor{amethyst}{rgb}{0.6, 0.4, 0.8}

\newcommand{\probP}{\mathbb{P}}
\newcommand{\x}{\bm{x}}

\newcommand{\btheta}{\bm{\theta}}
\newcommand{\bomega}{\bm{\omega}}
\newcommand{\cD}{\mathcal{D}}
\newcommand{\cX}{\mathcal{X}}
\newcommand{\cY}{\mathcal{Y}}

\newcommand{\cG}{\mathcal{G}}

\usepackage{amsmath,amsfonts,bm}

\def\eqref#1{equation~\ref{#1}}

\def\1{\bm{1}}

\DeclareMathAlphabet{\mathsfit}{\encodingdefault}{\sfdefault}{m}{sl}
\SetMathAlphabet{\mathsfit}{bold}{\encodingdefault}{\sfdefault}{bx}{n}

\DeclareMathOperator*{\argmin}{arg\,min}

\newcommand{\Rbb}{\ensuremath{\mathbb{R}}}

\newcommand{\inv}[1]{\ensuremath{\frac{1}{#1}}}

\renewcommand{\leq}{\ensuremath{\leqslant}}

\newcommand{\ma}[1]{\ensuremath{\mathsf{#1}}}
\renewcommand{\vec}[1]{\ensuremath{\bm{#1}}}

\newcommand{\norm}[1]{\ensuremath{\left\| #1\right\|}}
\newcommand{\abs}[1]{\ensuremath{\left| #1 \right|}}

\newcommand{\set}[1]{\ensuremath{\mathcal{#1}}}

\definecolor{better}{rgb}{0.19, 0.55, 0.91}
\definecolor{worse}{rgb}{0.82, 0.1, 0.26}
\definecolor{first}{rgb}{0.13, 0.67, 0.8}
\definecolor{second}{rgb}{0.74, 0.83, 0.9}
\definecolor{aeroblue}{rgb}{0.79,1.0,0.9}

\usepackage{hyperref}
\usepackage{url}

\title{Take One Gram of Neural Features,\\ Get Enhanced Group Robustness}

\author[*,1,3]{\bf{Simon Roburin}}
\author[*,2,3]{\bf{Charles Corbi{\`e}re}}
\author[3]{\bf{Gilles Puy}}
\author[2,4]{\bf{Nicolas Thome}}
\author[1]{\bf{Mathieu Aubry}}
\author[1,3]{\bf{Renaud Marlet}}
\author[3]{\bf{Patrick P{\'e}rez}}

\affil[1]{\footnotesize LIGM, \'Ecole des Ponts, France}
\affil[2]{\footnotesize CEDRIC, Conservatoire National des Arts et Métiers, Paris, France}
\affil[3]{\footnotesize valeo.ai, Paris, France}
\affil[4]{\footnotesize ISIR, Sorbonne University, Paris, France}

\affil[ ]{\texttt{\{simon.roburin, mathieu.aubry\}@enpc.fr}}
\affil[ ]{\texttt{\{charles.corbiere, nicolas.thome\}@cnam.fr}}
\affil[ ]{\texttt{\{gilles.puy, renaud.marlet, patrick.perez\}@valeo.com}}

\iclrfinalcopy 
\begin{document}

\maketitle
\def\thefootnote{*}\footnotetext{Equal contribution}\def\thefootnote{\arabic{footnote}}

\begin{abstract}
Predictive performance of machine learning models trained with empirical risk minimization (ERM) can degrade considerably under distribution shifts. In particular, the presence of spurious correlations in training datasets leads ERM-trained models to display high loss when evaluated on minority groups not presenting such correlations in test sets. Extensive attempts have been made to develop methods improving worst-group robustness. However, they require group information for each training input or at least, a validation set with group labels to tune their hyperparameters, which may be expensive to get or unknown a priori. In this paper, we address the challenge of improving group robustness \emph{without group annotations during training}. To this end, we propose to partition automatically the training dataset into groups based on Gram matrices of features extracted from an identification model and to apply robust optimization based on these pseudo-groups. In the realistic context where no group labels are available, our experiments show that our approach not only improves group robustness over ERM but also outperforms all recent baselines.
\end{abstract}

\section{Introduction}
Empirical Risk Minimization (ERM) is the most standard machine learning formulation, which assumes that training and testing samples are independent and identically distributed \citep{NIPS1991_ff4d5fbb}. 
While academic datasets are mainly built to respect this assumption, practical settings display more challenging configurations with distribution shifts. Among different types of shifts, training data can be affected by selection biases and confounding factors, also called \emph{spurious correlations} \citep{woodward2005making,duchi2019distributionally} 

Imagine crowd-sourcing an image dataset of camels and cows \citep{Beery2018RecognitionIT}. Due to selection biases, a large majority of cows stand in front of grass environment and camels in the desert. A simple way to differentiate cows from camels would be to classify the background, an undesirable shortcut that ERM will naturally exploit. Consequently, ERM may perform poorly on minority groups that do not display such spurious correlations~\citep{hashimoto2018fairness,tatman2017gender,duchi2019distributionally}, \eg, a cow standing in the desert. To overcome this issue, recent works \citep{creager21environment,bao2022, sohoni2021,liu2021,ahmed2021systematic,Kirichenko2022} 
rely on {two-stage} schemes: first, automatic environment discovery (\eg, based on deep feature clustering); then, robust optimization based on environment pseudo-labels. 
\emph{Environment} here refers to a recurring setting, not intrinsic to the object of interest, that may affect its classification, such as background, object color or object pose.
However, all these approaches require the availability of ground-truth environment labels on a validation set to properly tune their hyperparameters. 

\begin{figure}[t]
\centering
 \includegraphics[width=0.95\linewidth]{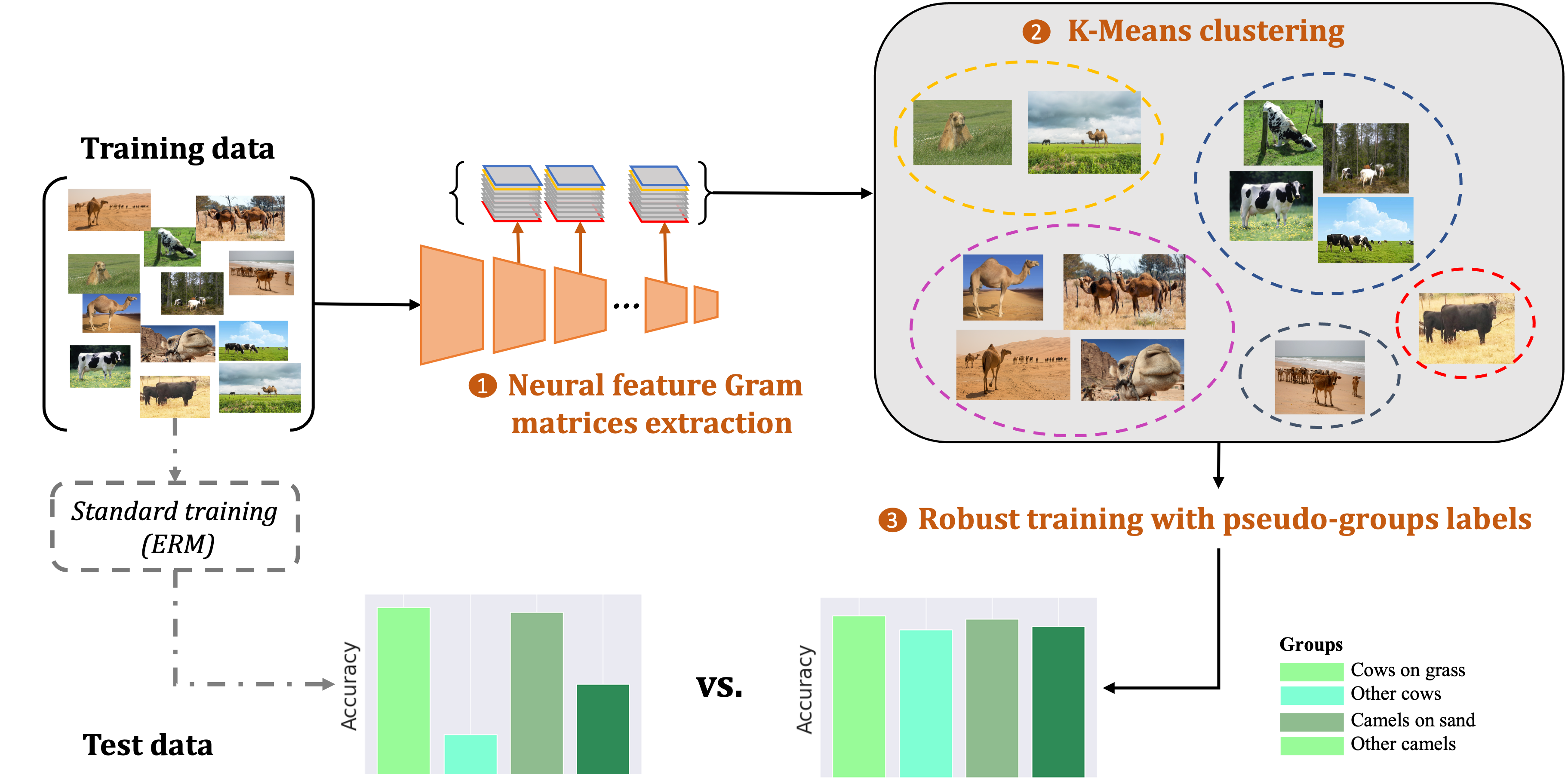}
 \caption{\textbf{Overview of the proposed \textsc{GramClust} approach for robust classification with unsupervised group discovery.}  (1)~We first extract deep image features using an identification model and (2)~we cluster the training dataset based on Gram matrices of images features; 
(3) Then, we train the targeted classifier with a robust optimization that exploits the assigned pseudo-group labels. Consequently, \textsc{GramClust} properly classifies samples in minority groups, \eg cows and camels in unusual environments -- in contrast to standard Empirical Risk Minimization (ERM) training.}
 \label{fig:intro}
\end{figure}

This paper addresses the problem of learning a robust classifier, which, for instance, would not confuse a cow standing in the desert with a camel although not given 
any annotation about grass or desert. In computer vision, many identified spurious correlations are closely related to visual aspects, such as background \citep{Beery2018RecognitionIT}, texture \citep{geirhos2018imagenettrained}, image style \citep{hendrycks2021many}, physic attributes \citep{Liu_2015_17414} or camera characteristics \citep{koh2021wilds}. In this work, we assume that relevant environment labels can be inferred from visual feature statistics, and demonstrate they lead to meaningful environments and robust classifiers for standard datasets used to evaluate robust classification. We propose a two-stage approach, \textsc{GramClust}, which first assigns a group label, \ie, a class-environment pair label, by partitioning a training dataset into clusters of images with similar visual statistics and then trains a robust classifier based on these pseudo-group labels. Our approach is summarized in \cref{fig:intro}.
We use Gram matrices as visual descriptive statistics, which are second-order moments of neural activation. Gram matrices are well known for displaying impressive results in style transfer techniques \citep{gatystransfer2016}, but more importantly for the interpretation of our approach, \cite{li2017demystifying} demonstrate that matching Gram matrices between two groups of images is equivalent to aligning the respective distribution of each group, minimizing the Maximum Mean Discrepancy.
Therefore, our method can be interpreted as grouping images into clusters of similar feature distributions that are sensible candidates for environments. 

Our main contributions are as follows: (1) We introduce an easy-to-scale method to split training images among distinct pseudo-environments, based on feature Gram matrices extracted by a specifically-trained identification model; (2) \textsc{GramClust} alleviates the need of ground-truth group labels altogether, even in the validation set, 
as hyperparameters are set based on validation performance computed from our pseudo-groups; (3) Extensive experiments on various image classification datasets with spurious correlations show that \textsc{GramClust} outperforms all recent baselines addressing robustness without group annotation. In particular, on the realistic large-scale CelebA dataset \citep{Liu_2015_17414}, we improve worst-group test accuracy by $+24.3$ points.

\section{Related Work}

Robustness to distribution shift \citep{Rusak2020a,hendrycks2020augmix,gulrajani2021in,geirhos2018imagenettrained} has recently been an increasingly popular topic among machine learning researchers. \citet{koh2021wilds} distinguish two types of distribution shifts: \emph{domain generalization}, where test samples come from a different distribution than training datasets, and \emph{subpopulation shift}, where train and test distributions overlap but their relative proportion differs. With subpopulation shifts, the goal is to perform well even on the minority group, also referred to as \emph{group robustness}. In this study, we focus on the latter form of distribution shift. 

\textbf{Group robustness with group annotations.} Recent approaches propose to leverage group annotations during training to improve group robustness. IRM \citep{arjovsky2020invariant} augments the standard ERM term with invariance penalties across data from different groups. \citet{ahmed2021systematic} promote, through a simple penalty, identical prediction behaviour across groups. Other works \citep{Sagawa2020Distributionally,zhang2020coping} minimize explicitly the worst-group loss during training. Finally, \citet{sagawa2020investigation} re-balance majority and minority groups via re-weighting and sub-sampling. 

\textbf{Group robustness without group annotations.} The focus here is a more realistic setting in which group annotations are not available on the training data. 
\citet{creager21environment} derive a group inference objective from a trained identification model that maximizes variability across environments, and is differentiable w.r.t a distribution over group assignments. \citet{liu2021} introduce a simple method in which environments are defined by images on which a trained identification model performs poorly. \citet{sohoni2021} propose an unsupervised clustering algorithm in the feature space of a trained identification model. But these methods require implicitly or explicitly a small validation set with ground-truth group annotation: \citet{liu2021} explicitly tunes its hyperparameters on a small validation set with group annotation while \citet{creager21environment} and \citet{sohoni2021} use best hyperparameters used in the GroupDRO paper \citep{Sagawa2020Distributionally} to minimize the worst-group accuracy on their inferred groups. These best hyperparameters were actually found using a validation set with true-group labels in the original study. 

\textbf{Gram matrices.} The original work of \citet{gatystransfer2016} demonstrated impressive results to generate images with the style of an existing image. The style of a first image is transferred to a second one by matching Gram matrices of features extracted by a convolutional neural network. \citet{Sastry2020} also used Gram matrices in out-of-distribution detection to identify an anomaly by comparing their values to the respective range observed over the training data. 
Interestingly, \citet{li2017demystifying} demonstrate a formal equivalence between matching Gram matrices of neural activations with an $L_2$ norm and the MMD with the second-order polynomial kernel. 
This shows that Gram matrices are also implicitly used in the process of distribution alignment between images.
This finding motivates our approach, which consists in discovering pseudo-groups using Gram matrices.

\section{\textsc{GramClust}: A Clustering Approach for Robust Optimization}

Our method, \textsc{GramClust}, consists of two main steps. First, we discover pseudo-environments among the images of a given dataset (see \cref{subsec:partition}). Second, we train a robust classifier that leverages the inferred pseudo-environment labels to reduce classification errors due to spurious environment correlations (see \cref{sec:rob_train}). To discover environments, we train during a few iterations an exogenous “identification model”. Then, using this model, we compute for each image its Gram matrix representation from different layers and apply random projections to reduce dimension. The resulting concatenated 
features are then fed to an unsupervised clustering algorithm ($k$-means) to produce pseudo-environment labels. 
This allows us to define pseudo-groups 
as the intersection of pseudo-environments 
and 
classes. 
Last, we train the target classifier by minimizing the standard cross-entropy classification loss on the worst pseudo-group with GroupDRO~\citep{Sagawa2020Distributionally}. 

\subsection{Problem formulation and notations}
Let us consider a dataset $\mathcal{D} = \{(\x_i, y_i)\}_{i=1}^N \in \cX \times \cY$ of $N$ samples where $\mathcal{X}$ is the input space and $\cY=\ \llbracket 1, K \rrbracket$ a set of labels.
We assume the data is sampled from random variables $(\bm{X}_e, \bm{Y}_e)$ in $\mathcal{X} \times \mathcal{Y}$ with probability law $\probP_{(\bm{X}_e, \bm{Y}_e)}$ for all $e \in  \llbracket 1, E \rrbracket$, where $E$ is the number of \textit{environments}. The full dataset can then be seen as the union of subsets associated to each random variable, i.e., $\mathcal{D} = \bigcup_{e=1}^E \mathcal{D}_e$ where each $\mathcal{D}_e$ is composed of \textit{i.i.d.}\ realisations of a random variable with joint probability law $\probP_{(\bm{X}_e, \bm{Y}_e)}$. 
For notation purposes, we actually choose the following equivalent formulation for the dataset $\mathcal{D} = \{(\x_i, y_i, e_i)\}_{i=1}^N \in \cX \times \cY \times \llbracket 1, E \rrbracket$ where $e_i$ refers to the environment from which $\x_i$ and $y_i$ were sampled. 

Our goal is to find a model $m$ in a given hypothesis space $\mathcal{M}$ that minimizes the error on the worst group. A \textit{group} is defined as a set of samples both from the same class and in the same environment. Formally, we introduce group distributions:
\begin{equation}
    \probP_{\bm{G}_{1,1}} = \probP(\bm{X}_1 | \bm{Y}_1 = 1),~\cdots, ~\probP_{\bm{G}_{E,K}} = \probP(\bm{X}_E | \bm{Y}_E = K).
\end{equation}

The purpose is then to solve the following objective minimization problem:
\begin{equation}
\label{eq:groupdro}
    \argmin_{m \in \mathcal{M}} \Big \{ \max_{g \in   \llbracket 1, E \rrbracket \times   \llbracket 1, K \rrbracket} \mathbb{E}_{(\x,y) \sim \probP_{\bm{G_g}}} \big [ \ell(m(\x),y) \big ] \Big \}, 
\end{equation}
    
where $\ell: \cY \times \cY \rightarrow \mathbb {R}^{+}$ is the cross-entropy loss between the model's prediction and the true label. 
Note that we have no access to any environment labels.
To circumvent this issue, we first discover pseudo-environment labels, then estimate the pseudo-group distributions to be used in \cref{eq:groupdro}. 

\subsection{Dataset partition}
\label{subsec:partition}

\begin{figure}[t]
\centering
 \includegraphics[width=0.9\linewidth]{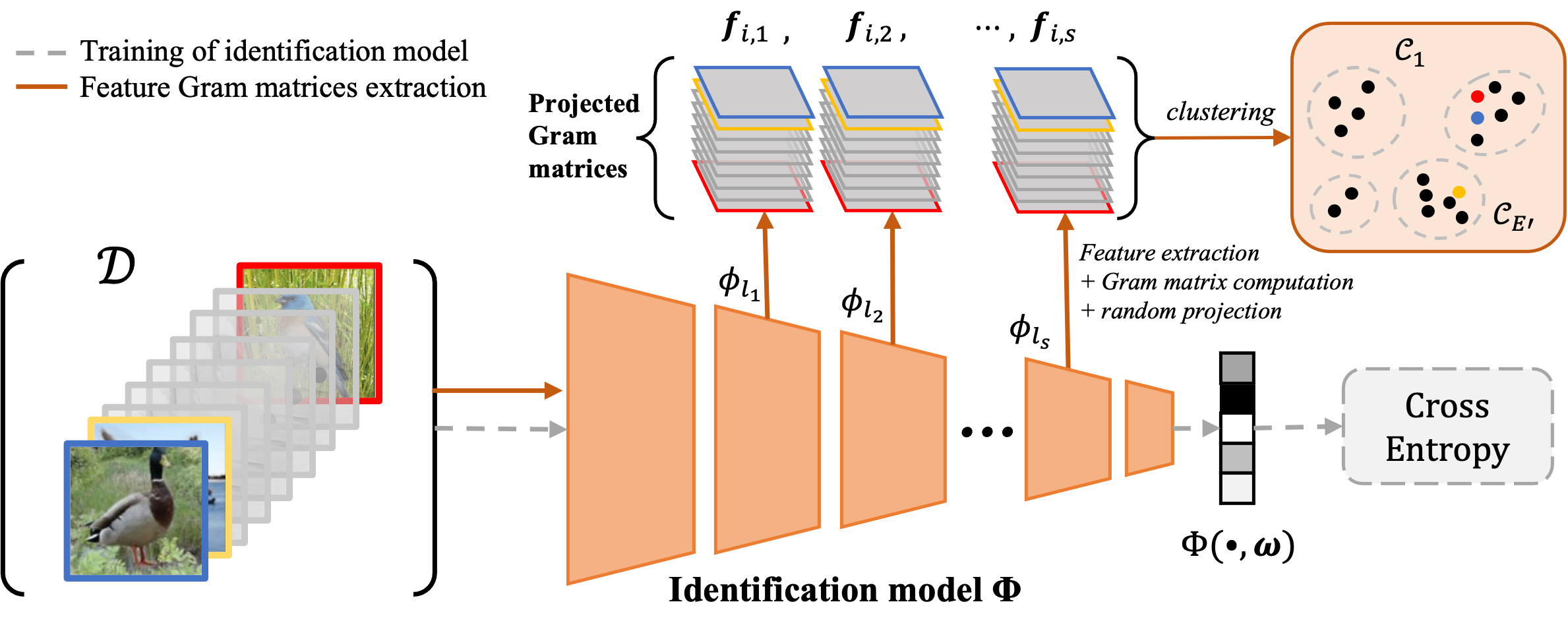}
 \caption{\textbf{Illustration of the proposed dataset partition}. An identification model $\mathrm{\Phi}$ with parameters $\bomega$ is trained for a limited number $T$ of epochs with ERM to fit groups with easy-to-learn spurious correlations. Then, for each image $\x_i \in \cX$, we extract intermediate features $\phi_l$ at layer~$l$ and compute their Gram matrix $\ma{G}_{l}$ with a random projection. These projected Gram matrix representations are used as features to cluster the training dataset $\cD_{\text{train}}$ in $E'$ environments.}
 \label{fig:schema_method}
\end{figure}

In this section, we describe the first stage of $\textsc{GramClust}$, which aims at environment discovery. The method is illustrated in \cref{fig:schema_method}.

\noindent \textbf{Identification model.}
Our approach starts by initializing a convolutional neural network $\mathrm{\Phi}$ for the classification task at hand; it is composed of $L$ layers with parameters $\bomega$ and is pre-trained on ImageNet \citep{deng2009imagenet}. \citet{liu2021} observed that ERM tends to fit models on data presenting easy-to-learn spurious correlations at the beginning of the learning process. It is only after a significant number of epochs that the model starts to learn more difficult patterns. Hence, we only train $\mathrm{\Phi}$ during a few iterations, minimizing w.r.t. $\bomega$ the following empirical loss function: 
\begin{equation}
    \frac{1}{N}\sum_{i=1}^{N} \ell(\mathrm{\Phi}(\x_i,\bomega),y_i),
\end{equation}
where $\ell: \cY \times \cY \rightarrow \mathbb{R}^{+}$ is the cross-entropy loss between the model's predicted label $\mathrm{\Phi}(\x_i,\bomega)$ and the true label $y$ associated with sample $\x$. 

In the following, we call $\mathrm{\Phi}$ the \emph{identification model} as our clustering is based on features extracted from this model. The idea is to leverage the biases learned by $\mathrm{\Phi}$ to identify relevant environments and partition the training dataset into groups of images presenting spurious correlations, on the one hand, and groups of images free from these correlations on the other hand. Hence, after this initial training and in the rest of the paper, the parameters $\bomega$ of the identification model $\mathrm{\Phi}$ are frozen.


\noindent \textbf{Features Gram matrices.} 
We denote the feature map of an image $\vec{x}$ at layer $l$ of $\mathrm{\Phi}$ by $\phi_l(\x) \in \Rbb^{M_l \times C_l}$, where $C_l$ is the number of channels and $M_l$ is the spatial size of the feature map. For each image $\x_i \in \cX$, we extract its feature maps at $S \leq L$ different and fixed layers $\{l_1, \cdots l_S\}$, and compute the Gram matrices defined as:
\begin{equation}
    \ma{G}_{l}(\x_i) = \inv{M_l} \; \phi_{l}(\x_i)^\intercal \phi_{l}(\x_i) \in \Rbb^{C_l \times C_l},\quad l \in \{l_1, \cdots l_S\}.
\label{eq:gram_matrix}
\end{equation}
Given input $\x_i$ and identification model $\mathrm{\Phi}$, the Gram matrix of its feature $\phi_l(\x_i)$ encodes visual correlations via an inner product between each pair of vectorized feature maps. In visual style transfer \citep{gatystransfer2016}, these Gram matrices have been shown to encode the ``style'' of an image, that is, loosely speaking, its textures and color palette, by contrast with its ``structure''.

\noindent \textbf{Clustering with $k$-means.} 
For each image $\x_i$, we vectorize and normalize its $S$ associated Gram matrices: $\vec{f}_{i,l} = {\rm vec}(\ma{G}_l(\x_i)) / \norm{{\rm vec}(\ma{G}_l(\x_i))}_2 \in \Rbb^{C_l^2}$. The normalization permits us to balance the contributions of the different Gram matrices in the clustering loss. 
Each image $\x_i$ is thus encoded by the vector $\vec{f}_{i} = \left[\vec{f}_{i,1}, \ldots, \vec{f}_{i,S} \right] \in \Rbb^C$, where $C = \sum_l C_l^2$. Relying on the assumption that environments can be inferred from visual feature statistics, we propose to discover $E'$ environments, $E'$ being a parameter as $E$ is unknown, by clustering the $N$ training images into $E'$ clusters $\{\set{C}_1, \ldots \set{C}_{E'}\}$ via $k$-means clustering, i.e., by computing a solution to:
\begin{equation}
    \label{eq:kmeans}
    \min_{\{\set{C}_1, \ldots \set{C}_{E'}\}} 
    \sum_{e=1}^{E'}
    \inv{2 \abs{\set{C}_e}}
    \sum_{\big \{i,j | \mathbf{x}_i, \mathbf{x}_j \in \set{C}_e \times \set{C}_e \big \}}
    \norm{\vec{f}_{i} - \vec{f}_{j}}_2^2,
\end{equation}
where $\norm{\vec{f}_{i} - \vec{f}_{j}}_2^2 = \sum_{l=1}^S \norm{\vec{f}_{i,l} - \vec{f}_{j,l}}_2^2$.


\noindent \textbf{Scaling with random projections.} 
Storing all these vectors and computing distances between them in a high-dimensional space is computationally and memory expensive on large datasets. We overcome this difficulty by projecting the vectors $\vec{f}_{i,l}$ in a lower-dimensional space as proposed by \citet{achlioptas03}. Given a size $\ell_0$, we build a matrix $\ma{P} \in \Rbb^{\ell_0 \times C}$ whose entries $\ma{P}_{mn}$ are the realisation of independent random variables: $\ma{P}_{mn} = 1$ or $\ma{P}_{mn} = -1$ with probability $1/2$. Then we compute
\begin{equation}
    \tilde{\vec{f}}_{i,l} = \inv{\sqrt{\ell_0}} \ma{P}\vec{f}_{i,l}
\end{equation}
and substitute $\tilde{\vec{f}}_{i,l}$ for $\vec{f}_{i,l}$ in \cref{eq:kmeans}. We justify this choice by the fact that this projection preserves the distances $\norm{\vec{f}_{i,l} - \vec{f}_{j,l}}_2^2$ involved in the $k$-means objective of \cref{eq:kmeans}. Indeed, let $\epsilon \in ]0, 1[$ and $\ell_0 \propto \log(N)$, then with high probability,\footnote{We let the reader refer to Theorem 1.1 in \citep{achlioptas03} for the exact expression of this probability as a function of $\epsilon$, $N$ and $\ell_0$.} 
\begin{equation}
(1 - \epsilon)\norm{\vec{f}_{i,l} - \vec{f}_{j,l}}_2 
\; \leq \;
\| \tilde{\vec{f}}_{i,l} - \tilde{\vec{f}}_{j,l} \|_2 
\; \leq \;
(1 + \epsilon) \norm{\vec{f}_{i,l} - \vec{f}_{j,l}}_2,
\end{equation}
for all $(i, j) \in \llbracket 1, N \rrbracket ^2$. In practice, we choose $\ell_0 = \lfloor 100 \log(N) \rfloor$ 
which yields dimensions for $\tilde{\vec{f}}_{i,l}$ much lower than typical values of $C_l$. We remark that this choice of projection is independent of all $\vec{f}_{i,l}$ and thus can be defined and fixed before any feature extraction.

\subsection{Robust optimization with pseudo-group labels}\label{sec:rob_train}


Given these estimated environments, we define their intersection with classes as ``pseudo-groups''. 
%
%
%
Formally, given the predicted environment $\hat{e}_i \in \llbracket 1, E' \rrbracket$, 
of image $i$, its pseud-group label is $\hat{g_i} = (\hat{e}_i, y_i) \in \llbracket 1, E' \rrbracket \times \llbracket 1, 
K \rrbracket$. 

Going back to \cref{eq:groupdro}, the distributions over the groups $\probP_{\bm{G}_{\hat{g}}}$ are estimated by
\begin{equation}
    \hat{\probP}_{\bm{G}_{\hat{g}}} = \delta(\cG(\hat{g}))\quad \quad \text{for all  } \hat{g} \in \llbracket 1, E' \rrbracket \times \llbracket 1, K \rrbracket,
\end{equation}
where $\delta$ is the Dirac distribution, and
\begin{align}
    \cG(1, 1) &= \{(\x_i, y_i), i \in \llbracket 1, N \rrbracket \; | \;y_i=1, x_i \in \set{C}_1 \} \\ &\cdots  \notag \\
    \cG(E', K) &= \{(\x_i, y_i), i \in \llbracket 1, N \rrbracket \; | \; y_i=K, x_i \in \set{C}_{E'} \}
\end{align}
are the sets of images and labels associated with the pseudo-group labels.

Each training point $\x_i \in \cX$ is now associated with a class label $y_i$ and a pseudo-group annotation $\hat{g_i}$. We train a robust classifier $h$ with parameters $\btheta$ by minimizing the worst-group risk on the training dataset \citep{Sagawa2020Distributionally}:
\begin{equation}
    \hat{\btheta} \in \argmin_{\btheta} \Big \{ \max_{\hat{g} \in \llbracket 1, E' \rrbracket \times \llbracket 1, 
K \rrbracket}~ \inv{\abs{\cG(\hat{g})}}\sum_{(\x,y) \in \cG(\hat{g})} \big [ \ell(h(\x,\btheta),y) \big ]  \Big \},
\end{equation}
where the loss $\ell: \cY \times \cY \rightarrow \mathbb{R}^{+}$ remains the cross-entropy between the predicted label $h(\x,\btheta)$ of the robust classifier and the true label $y$ associated with sample $\x$.

\subsection{Model selection via cross-validation on validation data}
\label{subsec:model_selection}

Setting relevant hyperparameters is important in optimization algorithms to ensure a proper convergence. Hyperparameters tuning is performed with cross-validation using a held-out subset of training data. With robust optimization, worst-group accuracy of the final classifier is the go-to metric for model selection. Previous approaches rely on \textit{true} group labels of the validation set to define and assess performance on the worst group. In contrast, we do not rely on such a prior information. We 
partition the validation set using the clusters found on the training set and we conduct cross-validation based on the resulting pseudo-groups. In our experiments, we observe that this type of model selection is effective to achieve proper group robustness.
 
\section{Experiments}

In this section, we evaluate the capacity of \textsc{GramClust} to improve group robustness on image classification datasets with spurious correlations. In \cref{subsec:comparative_results}, we empirically show that it outperforms other baselines addressing robustness without group annotation on three datasets. We then present in \cref{subsec:empirical_analysis} an empirical analysis of our approach, including the importance of using Gram matrices as visual features, the impact of the choice of layers to extract features from, and the impact of the number of clusters. 
The code is available with the supplementary material.

\subsection{Setup}
\label{subsec:exp_setup}

\textbf{Datasets}. We experiment with three image classification datasets on which previous works evaluate worst-group performance. \textbf{Waterbirds} \citep{Sagawa2020Distributionally} is a dataset composed of images combining bird photographs from the CUB dataset \citep{WelinderEtal2010} with background scenes taken from the Places365 dataset \citep{placesdataset}.  The target labels (``landbirds''/``waterbirds'') are spuriously correlated with the background images (``land''/``water''). The train set is composed of 4,795 images and the validation and test set are respectively composed of 1,199 and 2,897 images. \textbf{CelebA} \citep{Liu_2015_17414} is a celebrity face dataset with 202,599 images. \citet{Sagawa2020Distributionally} considered the task of classifying the hair color of the individual as ``blond'' or ``not blond''. The authors observed that there exists a spurious correlation between the hair color and the gender (``male'' or ``female'') of a person. In fact, in the dataset, only 2\% of blond people are male. We use the official train-val-test split from \citet{Liu_2015_17414}. \textbf{COCO-on-Places-224} is a dataset of 10 segmented MS COCO \citep{lin2014microsoft} objects superimposed on scenes from the Places365 dataset. The train set has 7,200 training images --- 800 images per category --- and validation and test sets are composed of 900 images --- 100 images per category. Unlike Waterbirds, multiple backgrounds have spurious correlations with the object classification \citep{ahmed2021systematic}. We rebuilt this dataset based on the code provided by \citet{ahmed2021systematic}\footnote{\url{https://github.com/Faruk-Ahmed/predictive_group_invariance}}  but with images resized to $224 \times 224$ (instead of $64 \times 64$ in the original paper, which considerably degrades visual features of objects and background). 

\textbf{Baselines}. We compare our approach against the standard \textbf{ERM} baseline and recent methods that aim at robust predictions across groups without the use of train group annotations: \textbf{EIIL}~\citep{creager21environment}, \textbf{GEORGE}~\citep{sohoni2021}, and \textbf{JTT}~\citep{liu2021}. We also include robust methods that use train group annotations: \textbf{IRM}~\citep{arjovsky2020invariant}, importance weighting and \textbf{GroupDRO}~\citep{Sagawa2020Distributionally}. The latter methods and ERM were already implemented and we took care to reproduce results for all methods. Our results with baselines are in line with those reported respectively in each original paper. Note that our approach and GroupDRO share the same robust optimization objective (\cref{eq:groupdro}). Hence, \textsc{GramClust} would boil down to GroupDRO if discovered pseudo-groups were to match exactly the ones annotated in the dataset. 

\textbf{Training details}.
All methods, including ours, 
use a ResNet-50 \citep{resnet2015} architecture pre-trained on ImageNet as the \emph{robust classifier}. Models are optimized using SGD with momentum. For GroupDRO and ERM, we use the hyperparameters reported by \citet{Sagawa2020Distributionally} on Waterbirds and CelebA datasets. Further training details are available in \cref{appx:implem_details}. Note that hyperparameters have been selected with the use of a validation set with group labels. Regarding our approach, we select a VGG-19 \citep{Simonyan15} architecture for the \emph{identification model} $\mathrm{\Phi}$. Although dating from 2015, VGG-19 is still the go-to architecture for applications involving Gram matrices such as image style transfer \citep{zhang2022exact,hollein2022stylemesh, xie2022artistic}.
Results of our approach include two types of model selection via cross-validation: (i)~based on a validation set with true-group annotations (\textsc{GramClust}\textit{-orig}), and (ii)~based on pseudo-group labels (\textsc{GramClust}\textit{-cv}) predicted by our clustering (see \cref{subsec:model_selection}).

\textbf{Metrics}. We report worst-group and average test accuracy for Waterbirds and CelebA datasets. On COCO-on-Places-224, we follow the evaluation protocol proposed by \citet{ahmed2021systematic} and report predictive performance on the in-distribution test set, which follows the same distribution as the training set, and on the systematically-shifted test set, where the spurious correlations have been removed and COCO objects are composed with uniformly-sampled random backgrounds.

\subsection{Comparative results}
\label{subsec:comparative_results}

\begin{table}[t]
    \centering
    \caption{\textbf{Comparative results on Waterbirds, CelebA and COCO-on-Places-224.} Worst-group (\textit{w-g}) and average (\textit{avg}) test accuracies (\% mean and std.)\ for Waterbirds and CelebA datasets; systematically-shifted (\textit{shift}) and in-distribution (\textit{in-dis}) test-set accuracies (\% mean and std.)  for COCO-on-Places-224 dataset. Experiments are with ResNet-50 models. \underline{Underlined} and \textbf{bold} type indicate respectively best and per-block best performance (with significance $p\,{<}\,0.05$ according to paired t-test on five runs). Methods are grouped according to their need for ground-truth group labels on train and/or val set; proposed \textsc{GramClust}\textit{-cv} is the only one requiring none.}
	\resizebox{\linewidth}{!}{%
    	\begin{tabular}{lcc|cc|cc|cc}
    		\toprule
    		 \multicolumn{3}{r|}{\textit{Group labels~}} & \multicolumn{2}{c|}{\textbf{Waterbirds}} & \multicolumn{2}{c|}{\textbf{CelebA}} & \multicolumn{2}{c}{\textbf{COCO-on-P}} \\
    		Method & train & val  & (w-g) & (avg) & (w-g) & (avg) & (shift) & (in-dis) \\
    		\midrule
    		ERM & & \checkmark & 65.0{\tiny $\pm 2.7$} ~& \underline{97.3}{\tiny $\pm 0.1$} & 42.4{\tiny $\pm 1.5$} ~& \underline{94.8}{\tiny $\pm 0.1$} & 71.9{\tiny $\pm 0.3$} ~& \underline{95.5}{\tiny $\pm 0.1$} \\
    		\midrule
    		IRM~\citep{arjovsky2020invariant} & \checkmark & \checkmark  & 77.4{\tiny $\pm 0.3$} ~& \underline{\ubold{97.3}}{\tiny $\pm 0.1$} & 75.1{\tiny $\pm 0.6$} ~& \ubold{94.5}{\tiny $\pm 0.1$} & 78.8{\tiny $\pm 0.3$} ~& \ubold{95.1}{\tiny $\pm 0.2$} \\
    		Importance weighting & \checkmark & \checkmark  & 74.4{\tiny $\pm 0.6$} ~& \underline{\ubold{97.4}}{\tiny $\pm 0.1$} & 72.4{\tiny $\pm 1.4$} ~& \ubold{94.4}{\tiny $\pm 0.2$} & 71.7{\tiny $\pm 0.5$} ~&  93.7{\tiny $\pm 0.2$} \\
    		GroupDRO~\citep{Sagawa2020Distributionally} & \checkmark & \checkmark & \ubold{83.9}{\tiny $\pm 0.3$} ~& 96.8{\tiny $\pm 0.1$} & \underline{\ubold{85.7}}{\tiny $\pm 2.0$} ~& 93.7{\tiny $\pm 0.2$} & \underline{\ubold{79.0}}{\tiny $\pm 0.4$} ~& 95.2{\tiny $\pm 0.2$}\\
    		\midrule
    		EIIL~\citep{creager21environment} & & \checkmark & 78.7{\tiny $\pm 0.3$} ~& \ubold{96.9}{\tiny $\pm 0.1$} & - ~& - & 68.5{\tiny $\pm 0.4$} ~& 94.8{\tiny $\pm 0.3$} \\
    		GEORGE~\citep{sohoni2021} & & \checkmark & 76.2{\tiny $\pm 2.0$} ~& 95.7{\tiny $\pm 0.5$} & 53.7{\tiny $\pm 1.3$} ~& \ubold{94.6}{\tiny $\pm 0.2$} & 71.6{\tiny $\pm 0.3$} ~& 95.1{\tiny $\pm 0.1$} \\
    		JTT\footnotemark~\citep{liu2021} & & \checkmark & 82.9{\tiny $\pm 0.3$} ~& 96.4{\tiny $\pm 0.2$} & 56.0{\tiny $\pm 0.7$} ~& 93.6{\tiny $\pm 0.0$} & 69.2{\tiny $\pm 0.4$} ~& 94.7{\tiny $\pm 0.3$} \\
    		\textsc{GramClust}\textit{-orig} (Ours) & & \checkmark & \underline{\ubold{85.3}}{\tiny $\pm 1.1$} ~& 96.6{\tiny $\pm 0.1$} & \ubold{77.9}{\tiny $\pm 2.2$} ~& 94.2{\tiny $\pm 0.2$} & \ubold{72.4}{\tiny $\pm 0.4$} ~& 95.0{\tiny $\pm 0.2$} \\
    		\midrule
    		\rowcolor{Gray} 
    		\textsc{GramClust}\textit{-cv} (Ours) & & & \underline{\ubold{85.3}}{\tiny $\pm 1.1$} ~& 96.6{\tiny $\pm 0.1$} & \ubold{80.3}{\tiny $\pm 1.9$} ~& 93.4{\tiny $\pm 0.1$} & \ubold{73.2}{\tiny $\pm 0.3$} ~& \ubold{95.3}{\tiny $\pm 0.3$} \\
    		\bottomrule
    	\end{tabular}
    }
    \label{tab:comparative_experiment}
\end{table}

We report quantitative comparisons on Waterbirds, CelebA and COCO-on-Places-224 
in \cref{tab:comparative_experiment}. We observe that \textsc{GramClust} improves worst-group test accuracy over ERM baseline on Waterbirds and CelebA and systematic generalisation on COCO-on-Places224. More importantly, \textsc{GramClust}\textit{-cv} achieves state-of-the-art performance on group robustness compared to all methods that do not use group labels on the training set. This results show empirically that our proposed approach, using Gram matrices of feature to discover pseudo-groups, which are then used for robust optimization and hyperparameter cross-validation, is very effective for group robustness. It also supports that Gram matrices are well suited to capture various types of dataset biases (background for Waterbirds, physical attribute in CelebA, multiple backgrounds in COCO-on-Places-224).
For instance, on Waterbirds, \textsc{GramClust}\textit{-cv} achieves $85.3\%$ worst-group accuracy compared to the second-best method, JTT, which reaches $82.9\%$. The gap is even more pronounced on CelebA where our approach outperforms JTT by $24.3$ pts. CelebA constitutes an interesting dataset to evaluate the scalability of methods as the training dataset is composed of ~200k images. For instance, we were not able to scale EIIL on this dataset.\footnotetext[3]{Results with JTT differ from the original paper as the scores that we report correspond to models trained without early-stopping.} Note that \textsc{GramClust}\textit{-orig} uses the same hyperparameters as EIIL, GEORGE and JTT for robust training of the target classifier from predicted group labels, and still displays significant improvements on the three datasets in terms of worst-group accuracy. \citet{liu2021} reported results that were obtained with early stopping thanks to a small validation set annotated with group labels. The authors selected models before convergence (around epoch 3) with low average accuracy on the test set but high worst-group accuracy. We argue that it is not a suitable property for a model and prefer models with high accuracy both in average and on the worst group of the test set.
Surprisingly, \textsc{GramClust}\textit{-cv} and \textsc{GramClust}\textit{-orig} outperform GroupDRO on Waterbirds with $85.3\%$ vs. $83.9\%$, while the latter method uses true-group labels during training. Our intuition is that it may be due to the ambiguity of the background in some Waterbirds images. We further discuss this result in \cref{subsec:illustration_waterbirds}. 
Overall, these results show that our pseudo-groups on the validation set are relevant to select good hyperparameters and more importantly, that \textsc{GramClust} does not require any group labels during training to achieve group robustness.

\subsection{Study of the clustering features}
\label{subsec:empirical_analysis}
In this section, we compare the performance obtained when clustering images with different visual features. In neural style transfer, \citet{huang2017adain} proposed the channel-wise mean and variance of image features, instead of Gram matrices as in \citep{gatystransfer2016}.
We thus compare the use of such features (\textit{`MeanVar'}) against our use of Gram matrices. 
We also compared our use of VGG-19 
features with the direct use of the penultimate representation of a ResNet-50 identification model (`{\it AvgPool}'). Recall that although our features for group identification are extracted using a VGG-19, our robust classifier is a ResNet-50. One may wonder if using directly the deepest features before the classification head in a ResNet-50 (\textit{`Standard'}) could be better than using VGG-19 features. For a fair comparison, we trained the robust classifier with the same hyperparameters for each method, which are consistent with those found by \citet{Sagawa2020Distributionally} with GroupDRO.

\begin{figure}[t]
\begin{minipage}[c]{\linewidth}
\centering
    \captionof{table}{\textbf{Study of the clustering features}. Results in worst-group (Waterbirds, CelebA) and systematically-shifted (COCO-on-P) test-set accuracies (\%). Gram matrices show to be the most effective type of information to obtain improved group robustness.}
	\resizebox{0.8\linewidth}{!}{%
    	\begin{tabular}{lllcccc}
    		\toprule
    		Visual features~~~ & Architecture~~~ & ~~Layer~~ & ~Waterbirds~ & ~CelebA~ & COCO-on-P \\
    		\midrule
    		Standard & ResNet-50 & \textit{AvgPool} & 76.2{\tiny $\pm 2.0$} & 53.7{\tiny $\pm 1.3$} & 71.6{\tiny $\pm 0.3$} \\
    		MeanVar & VGG-19 & \textit{Conv5\_1} & \ubold{85.3}{\tiny $\pm 1.2$} & 69.8{\tiny $\pm 1.0$} & 71.4{\tiny $\pm 0.5$}  \\
            Gram matrix & VGG-19 & \textit{Conv5\_1} & \ubold{85.3}{\tiny $\pm 1.1$} & \ubold{77.9}{\tiny $\pm 2.2$} & \ubold{72.4}{\tiny $\pm 0.4$} \\
    		\bottomrule
    	\end{tabular}
    }
    \label{tab:ablation_style_features}
\end{minipage}
\vspace{-0.2cm}
\begin{minipage}[c]{0.55\linewidth}
\centering
    \includegraphics[width=0.72\linewidth]{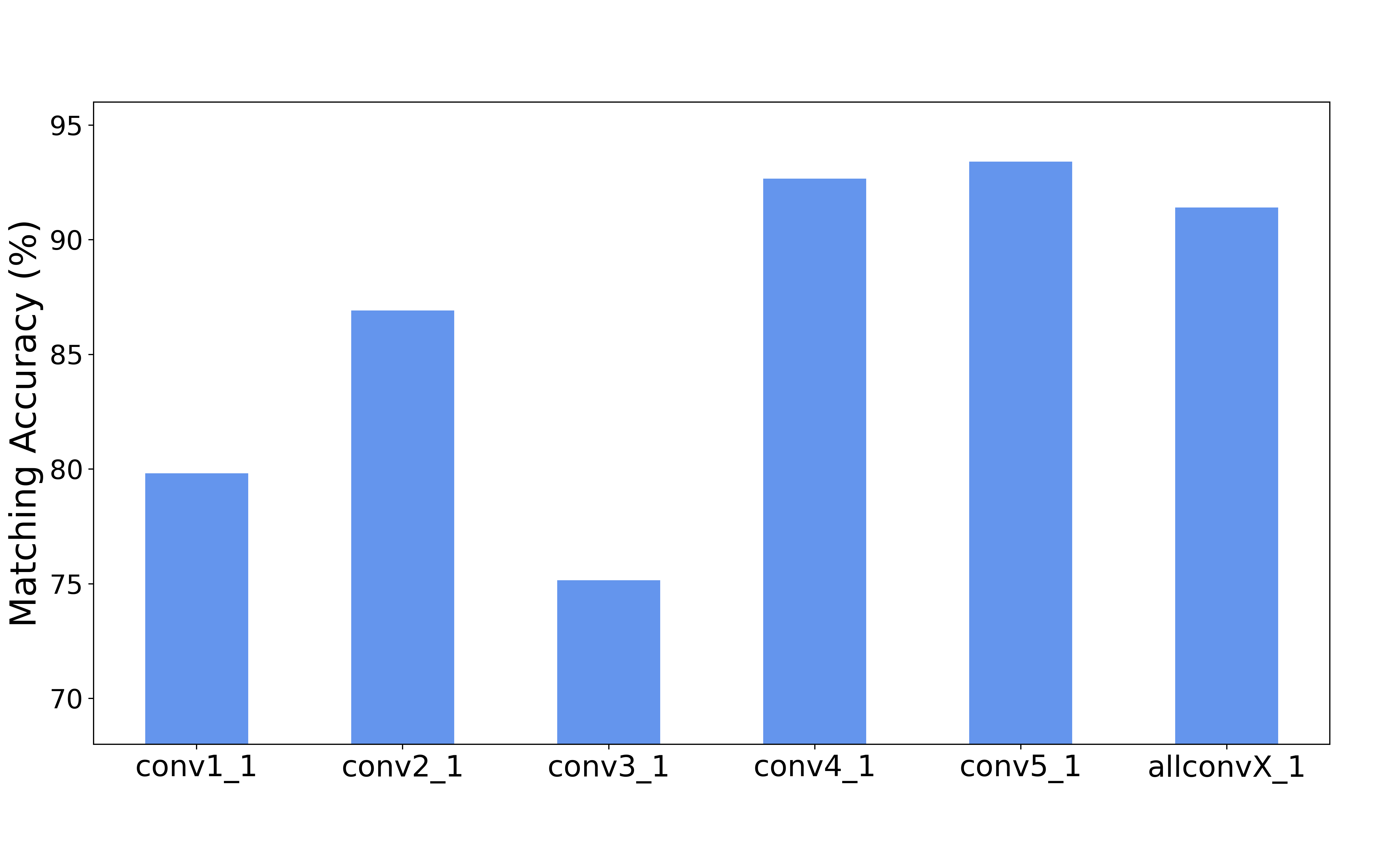}
    \vspace{-0.2cm}
    \caption{\textbf{Impact of the layer choice to extract features}. Results in matching accuracy on the validation set for \textsc{GramClust} on Waterbirds.}
    \label{fig:impact_choice_layer}
\end{minipage}%
\hspace{0.5cm}
\begin{minipage}[c]{0.41\linewidth}
\centering
    \includegraphics[width=0.61\linewidth]{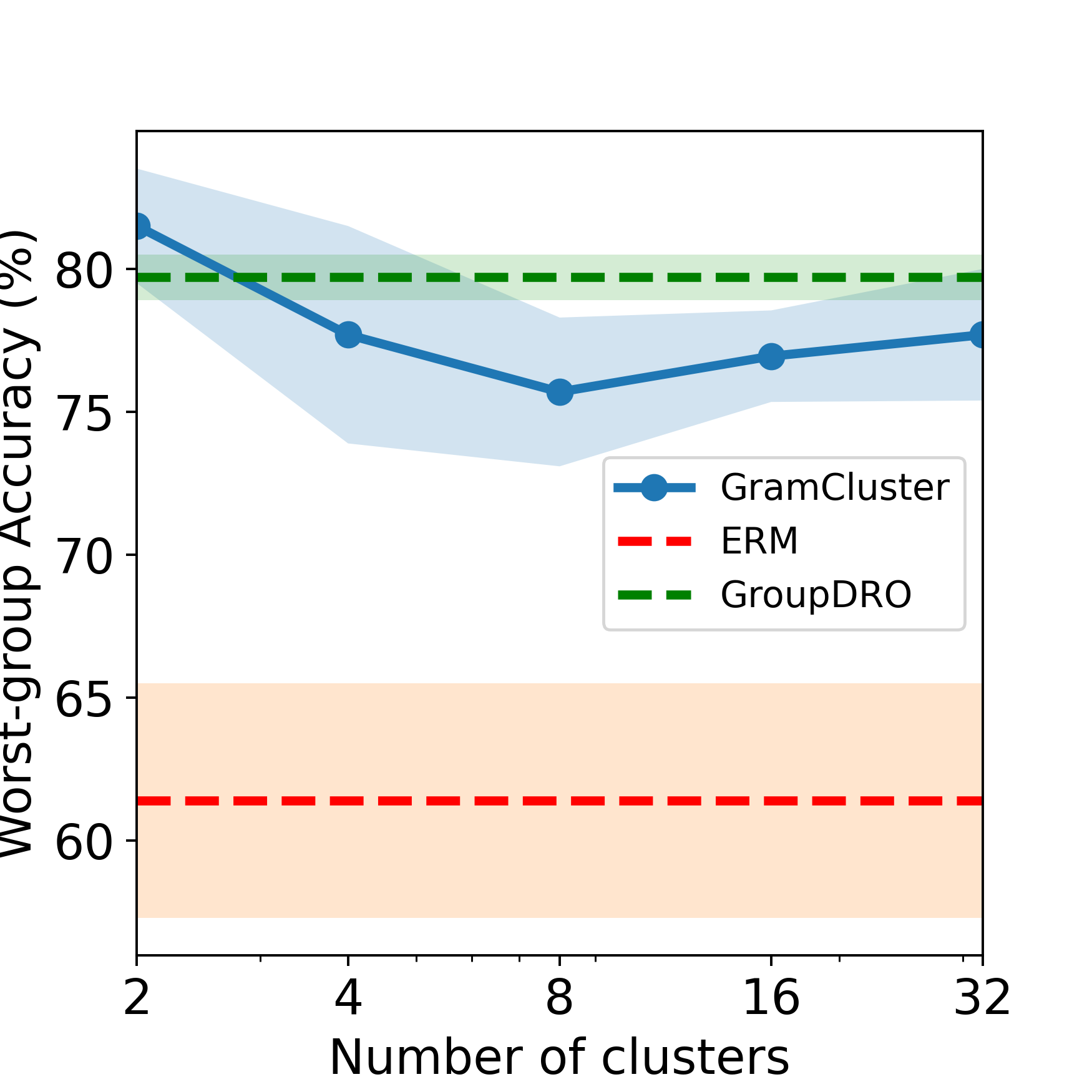}
    \vspace{-0.2cm}
    \caption{\textbf{Impact of the number of clusters.} Results in worst-group val accuracies of \textsc{GramClust} on Waterbirds.}
    \label{fig:impact_nb_clusters}
\end{minipage}
\end{figure}

The results are available in \cref{tab:ablation_style_features} for Waterbirds, CelebA and COCO-on-Places-224. Using the penultimate layer of a ResNet-50 as visual features for the clustering produces poorer performance than Gram matrices of VGG-19 features in every configuration. \textit{MeanVar} reaches test worst-group accuracy on-par with \textit{Gram matrix} on Waterbirds but degrades significantly performance on CelebA: $69.8\%$ in average compared to $77.9\%$ with \textit{Gram matrix}. Gram matrices provide more information than \textit{MeanVar} as their diagonals already contain the information about the channel-wise mean and variance of the deep features (see \cref{eq:gram_matrix}). This show that when scaling on large datasets such as CelebA, keeping all the correlations between different channels is important for group robustness.

\subsection{Clustering analysis}
\label{subsec:clustering_analysis}

\noindent \textbf{Effect of the selected layers for features.} We evaluate the impact of the selection of VGG-19 layers to extract the features in the clustering stage. To this end, we study the matching of the predicted environments to the true environment labels on the validation set. The assignment problem is solved via Hungarian matching \citep{Kuhn1955Hungarian} and we measure the global matching accuracy across all validation samples, where matching accuracy is the percentage of samples whose predicted group corresponds to its true group. In \cref{fig:impact_choice_layer}, we compare results on Waterbirds
using either one of the five layers commonly used in neural style transfer (\textit{conv1\_1}, \textit{conv2\_1}, \textit{conv3\_1}, \textit{conv4\_1}, \textit{conv5\_1}) or using all layers together.
Experiments show that: (i) Features from deeper layers correlate with better matching accuracy; (ii) Our approach is robust to the choice of deep layers either taken together (\textit{allconvX\_1}) or individually such as \textit{conv4\_1} and \textit{conv5\_1}; (iii) Using \textit{conv5\_1} outperforms selecting all traditional style layers and results in the highest score of $93.41\%$. We found consistent conclusions on the CelebA dataset. Results are reported in \cref{appx:clustering_celeba}.

\noindent \textbf{Impact of the number of clusters.} We study the impact of the number of clusters as hyperparameter in the clustering algorithm. Worst-group accuracy on the validation set for $E'\in \{ 2, 4, 8, 16, 32 \}$ clusters are reported in \cref{fig:impact_nb_clusters} for Waterbirds datasets. Overall, our method is robust to a variation in the number of clusters: \textsc{GramClust} with higher numbers of clusters produces a slight drop in performance but still outperforms ERM. It also has on-par performance with GroupDRO.

\subsection{Discussion about results of \textsc{GramClust} vs. GroupDRO on Waterbirds }
\label{subsec:illustration_waterbirds}

\begin{figure}[t]
\centering
\captionsetup[subfigure]{justification=centering}
\begin{minipage}[c]{0.48\linewidth}
\begin{minipage}[c]{0.30\linewidth}
\centering
    \includegraphics[width=\linewidth]{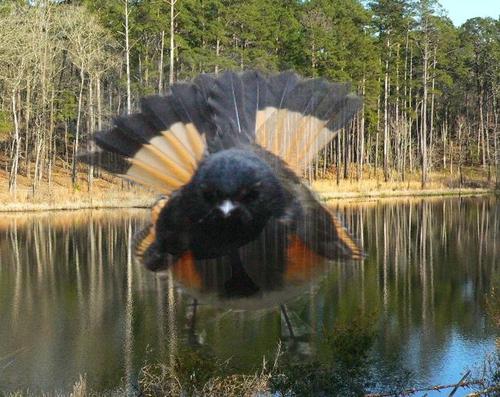}
\end{minipage}%
\hspace{0.1cm}
\begin{minipage}[c]{0.34\linewidth}
\centering
    \includegraphics[width=\linewidth]{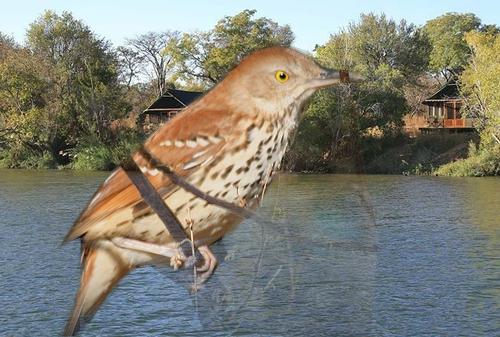}
\end{minipage}%
\hspace{0.1cm}
\begin{minipage}[c]{0.31\linewidth}
\centering
    \includegraphics[width=\linewidth]{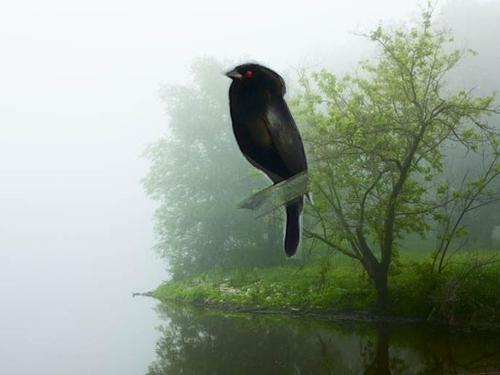}
\end{minipage}%
    \subcaption{True land background predicted as water background by \textsc{GramClust}}
\end{minipage}
\begin{minipage}[c]{0.48\linewidth}
\begin{minipage}[c]{0.39\linewidth}
\centering
    \includegraphics[width=\linewidth]{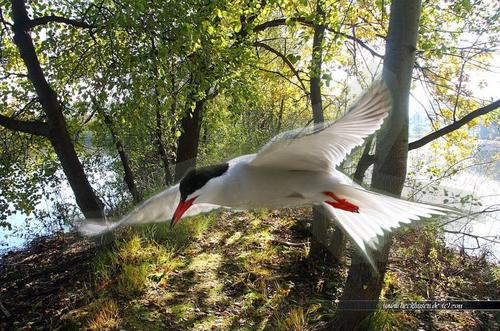}
\end{minipage}%
\hspace{0.1cm}
\begin{minipage}[c]{0.38\linewidth}
\centering
    \includegraphics[width=\linewidth]{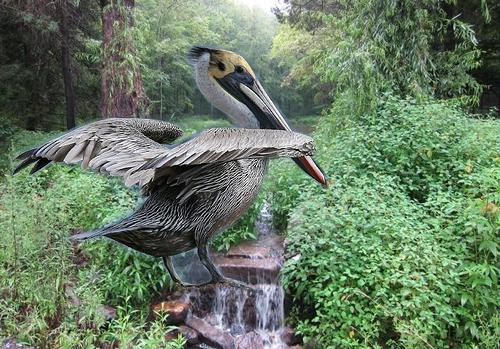}
\end{minipage}%
\hspace{0.1cm}
\begin{minipage}[c]{0.18\linewidth}
\centering
    \includegraphics[width=\linewidth]{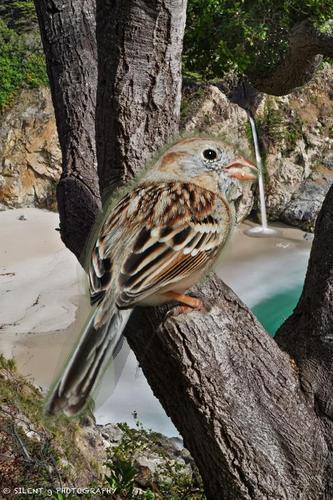}
\end{minipage}
\subcaption{True water background predicted as land background by \textsc{GramClust}}
\end{minipage}
\caption{\textbf{Example of confusing samples in Waterbirds dataset, wrongly predicted by \textsc{GramClust}.} (a) Samples of confusing land-background images predicted as water background; (b) Samples of confusing land-background images predicted as water background. In each case, the actual image background is confusing due to the joint presence of elements reflecting land background (forest, heavy vegetation, sand) and water background (water surface, rainfalls, mist).}
\label{fig:vizu_waterbirds}
\end{figure}

Comparative results in \cref{tab:comparative_experiment} show that \textsc{GramClust}\textit{-orig} outperforms GroupDRO on the Waterbirds dataset. The difference between the approaches lies in the usage of true-group labels on the training dataset for GroupDRO while \textsc{GramClust}\textit{-orig} leverages its predicted pseudo-groups. This results might be surprising given that the evaluation is performed on true test group labels and that the two methods
share the same robust optimization algorithm and hyperparameters. We intuit that this behavior, which occurs only on Waterbirds, is related to the group labels in the dataset. In \cref{fig:vizu_waterbirds}, we show some examples of confusing images that were not correctly assigned with our predicted group labels with \textsc{GramClust}. These images are taken from the set of mismatches between true-group labels and our pseudo-group labels after the Hungarian matching. We can see that some of these samples present dominant characteristic elements from land background, such as heavy vegetation and sand, while being labeled as water background. Conversely, some samples labeled as land background display a high percentage of water surfaces in the image. As mentioned in \cref{subsec:exp_setup}, the Waterbird dataset was created by combining bird photographs with background scenes taken from the Places365 dataset. But the latter dataset is composed of very diverse images which might not reflect the expected background for a category. 
This unwanted behavior outlines the difficulty of manually annotating groups and raises the need for creating benchmarks including datasets with spurious correlations from non-artificial, real-world data, such as hair color/gender bias observed in CelebA. 
It also motivates further research on the automatic discovery of groups in data, as proposed in our method.


\section{Conclusion}

In this paper, we introduce \textsc{GramClust}, a two-stage method that first partitions a training dataset into clusters via $k$-means clustering based on Gram matrices computed from image features, which are extracted from a identification model trained to catch spurious correlations in a biased dataset. This first stage is then followed by learning a robust classifier which minimizes the error on the worst pseudo-group labels previously discovered. \textsc{GramClust} demonstrates to be an effective approach to tackle group robustness and outperforms every baseline on standard datasets with spurious correlations. The usage of Gram matrices of features is crucial to capture pertinent visual statistics of the image and enables a relevant partition for robust training. Our approach also alleviates the need to label a validation set of images with group information and is able to tune its hyperparameters in an unsupervised fashion by applying its clustering algorithm on the validation set.

\bibliography{iclr2023_conference}
\bibliographystyle{iclr2023_conference}

\clearpage
\appendix

\section{Implementation details}
\label{appx:implem_details}

This section focuses on implementation details used to produce the results in the main text of our paper. The code that we used is provided along with this appendix. Our implementation builds upon the WILDS framework\footnote{\url{https://github.com/p-lambda/wilds}} released wby \citet{koh2021wilds}.

\subsection{Construction of COCO-on-Places-224}



We generated the dataset using the code\footnote{\url{https://github.com/Faruk-Ahmed/predictive_group_invariance}} of \citet{ahmed2021systematic} but, as explained in the main paper, we modified it to produce images of size $224 \times 224$ instead of $64 \times 64$. 
The reader can refer to the appendix of \citep{ahmed2021systematic} for more details regarding the generation of the COCO-on-Places dataset.

\subsection{Details about robust optimization}

We trained all models on one NVIDIA\textsuperscript{\textregistered} V100 Tensor Core with 16GB of memory, using PyTorch 1.10 and CUDA 10.2.

We used the implementations of IRM \citep{arjovsky2020invariant}, Importance Weighting and GroupDRO \citep{Sagawa2020Distributionally} available in WILDS \citep{koh2021wilds}, our own implementations of JTT \citep{liu2021} and of GEORGE \citep{sohoni2021} (while making sure that we could reproduce the original performance on Waterbirds and CelebA), and the official implementation\footnote{\url{https://github.com/ecreager/eiil}} of EIIL \citep{creager21environment}. Concerning EIIL, we recall that we were not able to make this method scale to large datasets such as CelebA.

For all methods, we used a ResNet-50 \citep{resnet2015} architecture trained using stochastic gradient descent with momentum (SGD-M) and $L_2$ regularization, but without any learning rate scheduler. We used a momentum of $0.9$ and a batch size of $128$ for all datasets and all methods. The learning rate $\eta$ and $L_2$ regularization parameters $\lambda$ are set as detailed below. 

JTT, GEORGE, EIIL, \textsc{GramClust} all use GroupDRO \citep{Sagawa2020Distributionally} as robust optimization step. On Waterbirds and CelebA, we did not redo any grid search and used the hyperparameters found in \citep{Sagawa2020Distributionally}. These hyperparameters were optimized using a small validation set annotated with true group labels. To produce the results on COCO-on-Places-224, we performed our own grid search using the annotated validation set. We considered values of $\eta$ and $\lambda$ close to those used in \citep{Sagawa2020Distributionally}: $\lambda \in \{10^{-4}, 10^{-2}, 10^{-1}, 1\}$ and $\eta \in \{10^{-5}, 5 \cdot 10^{-5}, 10^{-4}\}$. The best hyperparameters for GroupDRO are summarized in Table~\ref{tab:groupdro_hparam}.

To ensure fair comparisons, we also performed the same grid search over $\eta$ and $\lambda$ for ERM, IRM and Importance Weighting. The best hyperparameters for ERM and IRM are summarized for each dataset in Table~\ref{tab:erm_hparam} and Table~\ref{tab:irm_hparam}, respectively. Note that they correspond to those reported in \citep{Sagawa2020Distributionally} for Waterbirds and CelebA.

\begin{table}[ht]
    \setlength{\tabcolsep}{12pt}
    \centering
    \caption{\textbf{SGD-M hyperparameters for GroupDRO training}.}
	\resizebox{\linewidth}{!}{%
    	\begin{tabular}{lccccc}
    		\toprule
    		SGD-M hyperparameters & Waterbirds & CelebA & COCO-on-Places-224 \\
    		\midrule
    		Learning rate $\eta$ & $10^{-5}$ & $10^{-5}$ & $5 \cdot 10^{-5}$ \\
    		$L_2$ regularization $\lambda$ & $1.0$ & $0.1$ & $10^{-2}$ \\
    		\bottomrule
    	\end{tabular}
    }
    \label{tab:groupdro_hparam}
\end{table}

\begin{table}[ht]
    \centering
    \setlength{\tabcolsep}{12pt}
    \caption{\textbf{SGD-M hyperparameters for ERM training}.}
	\resizebox{\linewidth}{!}{%
    	\begin{tabular}{lccccc}
    		\toprule
    		SGD-M hyperparameters & Waterbirds & CelebA & COCO-on-Places-224\\
    		\midrule
    		Learning rate $\eta$ & $10^{-4}$ & $10^{-4}$ & $10^{-4}$ \\
    		$L_2$ regularization $\lambda$ & $10^{-3}$ & $10^{-4}$ & $10^{-4}$ \\
    		\bottomrule
    	\end{tabular}
    }
    \label{tab:erm_hparam}
\end{table}

\begin{table}[ht]
    \centering
    \setlength{\tabcolsep}{12pt}
    \caption{\textbf{SGD-M hyperparameters for IRM training}.}
	\resizebox{\linewidth}{!}{%
    	\begin{tabular}{lccccc}
    		\toprule
    		SGD-M hyperparameters & Waterbirds & CelebA & COCO-on-Places-224 \\
    		\midrule
    		Learning rate $\eta$ & $10^{-4}$ & $10^{-5}$ & $5 \cdot 10^{-5}$ \\
    		$L_2$ regularization $\lambda$ & $10^{-3}$ & $0.1$ & $0.1$ \\
    		\bottomrule
    	\end{tabular}
    }
    \label{tab:irm_hparam}
\end{table}

\subsection{Group discovery details}


For \textsc{GramClust}, we follow standard practice of neural style transfer \citep{gatystransfer2016} and use the VGG-19 \citep{Simonyan15} architecture for the identification model. This model is trained during 1 epoch on the training dataset with ERM using a batch size of $128$ and SGD-M. 
Among usual layers used to compute representations in neural style transfer, we observed improved performance by selecting deeper layers in the network (see \cref{subsec:clustering_analysis}). Consequently, for each dataset, we consistently extract features from the \textit{conv5\_1} layer, i.e., the first convolutional layer of block $5$.
Following results of \cref{fig:impact_nb_clusters}, we set the number of clusters to $2$ in our experiments.

For EIIL and GEORGE, the identification model is a ResNet-50 as used in the original methods. We train the model for 1 epoch with ERM using SGD-M, as for \textsc{GramClust}. 
Note that the activation at the output of the last layer is a sigmoid in EIIL while it is a softmax in GEORGE. 
As for \textsc{GramClust}, the best results were obtained when using $2$ clusters for EIIL and GEORGE. We refer the reader to \citep{creager21environment} and \citep{sohoni2021} for other implementation details specific to  EIIL and GEORGE, respectively. 


\subsection{Cross validation on pseudo-group annotations}

We report in Table~\ref{tab:val_pseudogroup} the results of our grid search on the validation set of each dataset using the \emph{pseudo-annotations} discovered with our method, i.e., using our discovered environments instead of the ground-truth ones. Hence, the average and worst group accuracies in Table~\ref{tab:val_pseudogroup} are computed using the discovered pseudo-groups. The hyperparameters used in \textsc{GramClust}\textit{-cv} correspond to those which yield the best worst-group accuracy in this table.

\begin{table}[ht]
    \centering
    \caption{\textbf{Grid search results on the validation sets of Waterbirds, CelebA and COCO-on-Places-224 with pseudo-group labels.} We report the worst-group (`{w-g}') and average (`{avg}') accuracies for Waterbirds and CelebA datasets, and the systematically-shifted (`{shift}') and in-distribution (`{ind}') accuracies for COCO-on-Places (`COCO-on-P') dataset.}
	\resizebox{1.0\linewidth}{!}{%
	   \setlength{\tabcolsep}{5pt}
    	\begin{tabular}{llc|cc|cc|cc}
    		\toprule
    		 \multicolumn{3}{r|}{Hyperparam.} & \multicolumn{2}{c|}{Waterbirds} & \multicolumn{2}{c|}{CelebA} & \multicolumn{2}{c}{COCO-on-P} \\
    		Method & $~\lambda$ & $\eta$ & w-g & avg & w-g & avg & sys & ind \\
    		\midrule
    		\textsc{GramClust}\textit{-cv}~  & 0.01 & $1 \cdot 10^{-5}$ & 74.6 ~& 82.4 &  \ubold{86.0}
    		~& 93.2
    		& 62.8 ~& 92.3 \\
            & 0.01 & $5 \cdot 10^{-5}$ & 69.2 ~& 79.9 & 53.5
            ~& \ubold{94.6}
            & 70.7 ~& 76.5 \\
            & 0.01 & $1 \cdot 10^{-4}$ & 70.0 ~& 80.6 & - ~& -  & 78.5 ~& 82.7 \\
            & 0.1 & $1 \cdot 10^{-5}$ & 75.4 ~& 82.6 &  85.6
            ~& 93.7
            & \ubold{78.7} ~& 83.3 \\
            & 0.1 & $5 \cdot 10^{-5}$ & 73.8 ~& 82.4 &  85.0
            ~& 89.1
            & 70.4 ~& 76.4 \\
            & 0.1 & $1 \cdot 10^{-4}$ & 76.9 ~& 85.8 &  - ~& - & 76.2 ~& 81.2 \\
            & 1 & $1 \cdot 10^{-5}$ & \ubold{80.8} ~& \ubold{86.4} & - ~& - & 65.5 ~& 72.6 \\
            & 1 & $5 \cdot 10^{-5}$ & 0.0 ~& 23.1 & - ~& - & 0.1 ~& 11.1 \\
            & 1 & $1 \cdot 10^{-4}$ & 0.0 ~& 23.1 & - ~& - & 0.2 ~& 11.1 \\
    		\bottomrule
    	\end{tabular}
    }
    \label{tab:val_pseudogroup}
\end{table}

\section{Clustering analysis on CelebA}
\label{appx:clustering_celeba}


 
\begin{figure}[ht]
\centering
    \includegraphics[width=0.9\linewidth]{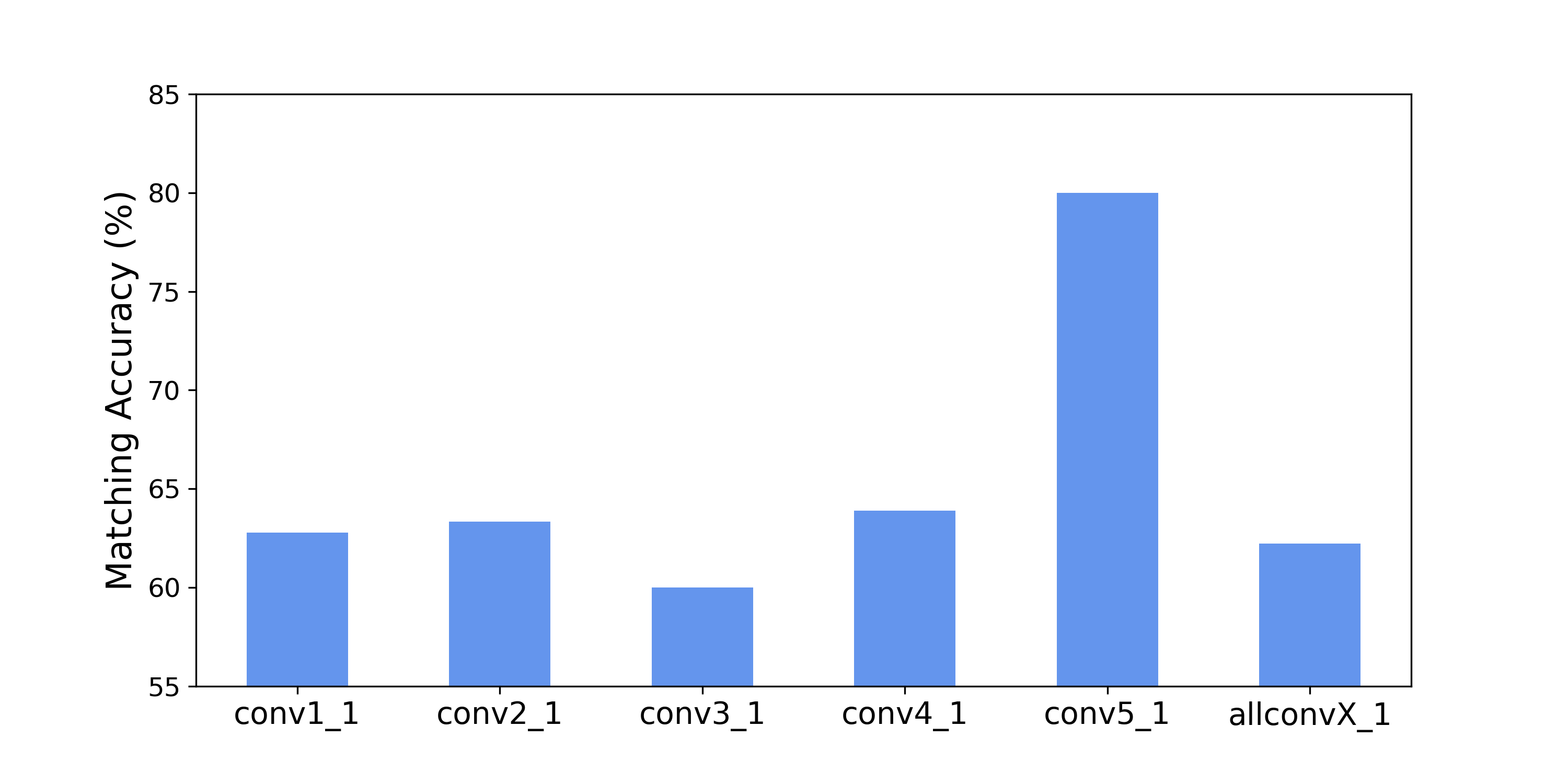}
    \caption{\textbf{Impact of the layer choice to extract features on CelebA}. We show the matching accuracy between the ground-truth environments on the validation set CelebA and the discovered ones with \textsc{GramClust} when using different VGG-19 layers. The result denoted $\textit{allconvX\_1}$ is obtained when using all the layers $\textit{conv1\_1}, \textit{conv2\_1}, \textit{conv3\_1}, \textit{conv4\_1}, \textit{conv5\_1}$ in our method.}
    \label{fig:celeba_layer}
\hspace{0.2cm}
\end{figure}

We present, in Figure~\ref{fig:celeba_layer}, the matching accuracy between the ground-truth environments and the environments discovered with our method on the validation set of CelebA for different layers of the VGG-19. As on Waterbirds, we notice that the best result is obtained when using the layer $\textit{conv5\_1}$.

\end{document}